\pdfoutput=1

\documentclass[11pt]{article}

\usepackage[final]{latex/acl}

\usepackage{times}
\usepackage{latexsym}

\usepackage[T1]{fontenc}

\usepackage[utf8]{inputenc}

\usepackage{microtype}

\usepackage{inconsolata}
\usepackage{amsmath}
\usepackage{arydshln}
\usepackage{siunitx}
\usepackage{graphicx}
\usepackage{booktabs}
\usepackage{balance}
\usepackage{colortbl}
\usepackage{enumitem}
\usepackage{multirow}
\usepackage{array}

\newcolumntype{C}[1]{>{\centering\arraybackslash}p{#1}}

\NewDocumentCommand{\zixuan}{ mO{} }{\textcolor{blue}{\textsuperscript{\textit{Zixuan}}\textsf{\textbf{\small[#1]}}}}

\NewDocumentCommand{\revanth}{ mO{} }{\textcolor{brown}{\textsuperscript{\textit{Revanth}}\textsf{\textbf{\small[#1]}}}}

\NewDocumentCommand{\heng}{ mO{} }{\textcolor{red}{\textsuperscript{\textit{Heng}}\textsf{\textbf{\small[#1]}}}}

\NewDocumentCommand{\kevin}{ mO{} }{\textcolor{green}{\textsuperscript{\textit{Kevin}}\textsf{\textbf{\small[#1]}}}}

\title{Towards Better Generalization in Open-Domain Question Answering by Mitigating Context Memorization}

\author{
  Zixuan Zhang$^{1}$, \textbf{Revanth Gangi Reddy}$^{1}$\textbf{,}  \textbf{Kevin Small}$^{2}$\textbf{,} \textbf{Tong Zhang}$^{1}$\textbf{,} \textbf{Heng Ji}$^{2}$ \\
  $^1$University of Illinois Urbana-Champaign, 
  $^2$Amazon,\\ 
  \texttt{\{zixuan11, revanth3, tozhang\}@illinois.edu}, \\
  \texttt{\{smakevin, jihj\}@amazon.com}
  }

\begin{document}
\maketitle
\begin{abstract}
Open-domain Question Answering (OpenQA) aims at answering factual questions with an external large-scale knowledge corpus. 
However, real-world knowledge is not static; it updates and evolves continually.
Such a dynamic characteristic of knowledge poses a vital challenge for these models, as the trained models need to constantly adapt to the latest information to make sure that the answers remain accurate.
In addition, it is still unclear how well an OpenQA model can transfer to completely new knowledge domains.
In this paper, we investigate the generalization performance of a retrieval-augmented QA model in two specific scenarios: 1) adapting to updated versions of the same knowledge corpus;  2) switching to completely different knowledge domains.
We observe that the generalization challenges of OpenQA models stem from the reader's over-reliance on memorizing the knowledge from the external corpus, which hinders the model from generalizing to a new knowledge corpus.
We introduce \emph{Corpus-Invariant Tuning} (CIT), a simple but effective training strategy, to mitigate the knowledge over-memorization by controlling the likelihood of retrieved contexts during training.
Extensive experimental results on multiple OpenQA benchmarks show that CIT achieves significantly better generalizability without compromising the model's performance in its original corpus and domain.
\end{abstract}

\section{Introduction}\label{sec:intro}

Open-domain Question Answering (OpenQA) \cite{chen-yih-2020-open} aims at answering factual questions using a large-scale external knowledge corpus. This is in contrast to closed-book question answering~\cite{roberts2020much} wherein the model is expected to directly answer questions with no access to external knowledge. 
In general, closed-book QA optimizes for memorization of knowledge in model parameters, while OpenQA focuses on retrieving relevant knowledge from an external corpus. 
OpenQA typically employs a retrieval-augmented approach~\cite{karpukhin-etal-2020-dense,iterative_training,em}, involving a two-stage process: a \emph{retriever} to select relevant documents, followed by a \emph{reader} to derive answers from these documents.
It is more practical for real-world applications as it enables the use of extensive and varied knowledge sources for answering questions.

Retrieval-augmented OpenQA models rely on an external corpus to physically store the knowledge.
However, real-world knowledge is not static; it updates and evolves continually.
Therefore, it is essential to build models that are able to use fresh and real-time knowledge~\cite{kasai2022realtime, vu2023freshllms}, but the dynamic characteristic of knowledge poses a vital challenge as the trained models need to constantly adapt to the latest information to make sure that the answers remain relevant and accurate. 
In addition, closed-book QA systems have been proved limited in adapting to new information or domains due to their reliance on pre-existing knowledge, and updating their parametric knowledge requires extensive large-scale pre-training. 
Nevertheless, it is still unclear how well OpenQA systems can transfer to leveraging unseen corpora and domains during training.

In this paper, we first investigate how well state-of-the-art retrieval-augmented models, such as  Atlas~\cite{atlas}, can adapt to new and diverse knowledge corpora. 
Specifically, we explore the model's performance in two scenarios: 1) adapting to updated versions of the same corpus (in \S{\ref{sec:RQ1}});  2) switching to completely different domains (in \S{\ref{sec:RQ2}}). Our investigation involves three settings: directly applying a pre-trained model, fine-tuning the model with the new corpora, and training the model afresh on the new corpora.
Initial experiments reveal that the model faces challenges in both scenarios. When directly transitioning to an updated corpus, there is a noticeable performance decline. Even additional tuning on the newer version doesn't achieve the same effectiveness as training from scratch with the new data (56.9$\rightarrow$59.5 vs 62.2 in Table \ref{tab:rq1_init_results}). Similar outcomes are observed when shifting from a general domain, like Wikipedia, to a specialized one, such as biomedical (41.2$\rightarrow$68.8 vs 69.7 in Table \ref{tab:rq2_init_results}).

We hypothesize and validate that such generalization challenges stem from the reader's over-reliance on memorizing the knowledge retrieved from the external corpus. This reliance primarily arises as the reader, with its primary training objective optimized for QA accuracy, often opts to hard-code a substantial amount of retrieved knowledge into its parametric memory. 
Such kind of over-memorization reduces the reader's dependency on the retriever to choose more relevant contexts. 
This phenomenon hampers the model's generalizability, particularly to updates in the knowledge corpus or changes in the knowledge domain. 
For instance, given a question \emph{Who is the prime minister of the UK?}, if a model has already hard-coded an outdated answer \emph{Boris Johnson} into its parameters (while being trained on an old corpus), it is harder to change its response even if the new information \emph{Rishi Sunak} from an updated corpus is available.

To address this issue, we introduce \emph{Corpus-Invariant Tuning} (CIT), a simple but effective training strategy to improve the corpus generalizability of retrieval-augmented text generation models. 
CIT aims to mitigate the reader's tendency to memorize the documents retrieved from the corpus during training. 
This pushes the reader to rely more on retrieved documents to answer the input questions, rather than relying on memorizing the knowledge facts into its parameters.
To achieve this, we propose a novel loss term to prevent memorization during training by controlling the likelihood of the retrieved documents.
Through extensive experiments across various OpenQA benchmarks~\cite{nq, tqa, han-etal-2023-robustqa}, carried out in both zero-shot and continual fine-tuning scenarios, we demonstrate that a retrieval-augmented model trained using our proposed CIT loss exhibits considerably enhanced generalizability across different corpora. This is evident by the considerable improvements in exact match (EM) scores, reaching up to a 2.1\% absolute gain.

Our contributions can be summarized as follows:
\begin{itemize}
    \item We propose to mitigate knowledge over-memorization of the reader to improve the generalization ability of retrieval-augmented text generation models.
    \item We introduce \emph{Corpus-Invariant Tuning} (CIT), a straightforward but effective training strategy that regularizes the reader's likelihood of the retrieved documents to mitigate it from over-memorizing the corpus during training.
    \item Through extensive experiments on multiple benchmarks, we demonstrate that training models with CIT greatly improves the generalization of OpenQA models across both newer versions of the corpora and unseen domains.
\end{itemize}

\section{Preliminaries}
\subsection{Problem Formulation}
Open-domain QA aims to answer questions only using a large-scale unified corpus, where the background documents for each question is not specified in advance. 
Given a natural language question $x$, our objective is to build a model $f(\cdot)$ to predict an answer $\hat{y}$ using a unified list of background documents $Z$, where $\hat{y}=f(x, Z)$.
Such a setting is more practical for real-world applications because it mirrors the vast and unstructured nature of real-world knowledge.
\paragraph{Retrieval Augmentation}
Since the external corpus collectively stores all essential information for answering the questions, the typical strategy to tackle the OpenQA problem is to implement a retrieval-augmented approach with a two-stage framework:
1) a \emph{retriever} to select a small subset of documents that are most relevant to the current question, and 2) a \emph{reader} to seek for useful information from the retrieved documents and generate the answer. 
Specifically, the probability of a predicted answer~$\hat{y}$ is decomposed by
\begin{equation*}
    p\left(\hat{y}\mid x, Z\right)=\sum_{\mathcal{C}\subset Z}p\left(\mathcal{C}\mid x;\theta\right)\cdot p\left(\hat{y}\mid \mathcal{C}, x;\phi\right),
\end{equation*}
where $\mathcal{C}$ denotes the set of retrieved documents, and $\theta$ and $\phi$ are the parameters of the retriever and the reader respectively.
During training, the retriever ($\theta$) and the reader ($\phi$) are often jointly optimized to ensure their effective collaboration, where the optimization is typically conducted with iterative training~\cite{iterative_training} or Expectation-Maximization (EM) based approach to train the model by treating the retrieved documents as hidden variables~\cite{em}.

\subsection{Evaluation of Model Generalization}\label{sec:pre_expts} 
We aim to tackle the generalization challenge for retrieval-augmented models as discussed in Section~\ref{sec:intro}.
Specifically, we focus on the following two main research questions (\emph{RQs}):
\begin{itemize}
    \item \emph{RQ1: How to improve the model's generalization ability across different versions (temporal snapshots) of the same corpus?}
    \item \emph{RQ2: How to improve the model's generalization ability across the corpora in different domains?}
\end{itemize}

\subsubsection{Evaluations of RQ1}
\label{sec:RQ1}
We conduct proof-of-concept experiments to test whether current retrieval-augmented OpenQA models can remain effective when the external corpus is updated to a newer version.
Specifically, we adopt the most recent retrieval-augmented model Atlas-XL~\cite{atlas} and test it on the Natural Questions (NQ) benchmark\footnote{\url{https://ai.google.com/research/NaturalQuestions}} with two different versions of Wikipedia (Wiki-2017 and Wiki-2018)\footnote{The Wikipedia dumped in 2017 and 2018 respectively.} as the external corpus.
We first fine-tune the Atlas-XL model on each version of Wikipedia, and then evaluate the model's generalization ability by both \emph{zero-shot testing} (train the model with Wiki-2017 and directly test it with Wiki-2018) and \emph{continue fine-tuning} (train the model with Wiki-2017 and further fine-tune it with Wiki-2018).
As shown in Table~\ref{tab:rq1_init_results}, we can first observe that the model performs better when initially fine-tuned with Wiki-2018, which shows that the updated KB can improve the performance.\footnote{The NQ benchmark is annotated in 2018, so Wiki-2018 is a more up-to-date background KB for the task.}
However, we can also observe a significant performance degradation when using the model trained with Wiki-2017 to directly test it on Wiki-2018.
Despite subsequent fine-tuning efforts, the performance still falls short of the original results obtained from initially training and testing with Wiki-2018.
These results indicate that the current retrieval-augmented models still struggle to effectively generalize when the background corpus undergoes evolution or updates.
\begin{table}[htbp]
	\centering
        \small
	\begin{tabular}{ccc}
		\toprule[1pt] 
            
		  Training & Testing & EM \\
		  Corpus & Corpus & Score~ \\
		\midrule[1pt]
		
		Wiki-2017  &   Wiki-2018 & 56.9 \\
            Wiki-2017 $\rightarrow$ Wiki-2018 &   Wiki-2018 & 59.5 \\
            \specialrule{0em}{1pt}{1pt}
        \cdashline{1-3}
        \specialrule{0em}{1pt}{1pt}
        Wiki-2018  & Wiki-2018 & 62.2 \\
		\midrule[1pt]
	\end{tabular}
	\caption{Initial experiments with Atlas-XL on the NQ benchmark with different Wikipedia versions. Results are evaluated with the exact-match (EM) score (\%).}
	\normalsize
 \label{tab:rq1_init_results}
\end{table}

\subsubsection{Evaluations of RQ2} 
\label{sec:RQ2}
We conduct similar experiments with Atlas-XL to evaluate its generalization ability across different domains.
We train the model on NQ with Wiki-2018 in the general domain, and test it on the \emph{Biomedical} split in RobustQA\footnote{\url{https://github.com/rujunhan/RobustQA-data}} with PubMed in the biomedical domain.
The results presented in Table~\ref{tab:rq2_init_results} reveal similar performance declines in both zero-shot and continual-fine-tuning settings, which indicates that the current OpenQA models also have inherent difficulty in generalizing across different domains.
\begin{table}[htbp]
	\centering
        \small
	\begin{tabular}{ccc}
		\toprule[1pt] 
            
		  Training & Testing & EM \\
		  Corpus & Corpus & Score \\
		\midrule[1pt]
            
		Wiki2018(NQ)  &   PubMed(Bio) & 41.2 \\
            Wiki2018(NQ)$\rightarrow$PubMed(Bio) &   PubMed(Bio) & 68.8 \\
         \specialrule{0em}{1pt}{1pt}
        \cdashline{1-3}
        \specialrule{0em}{1pt}{1pt}
          PubMed(Bio)  & PubMed(Bio) & 69.7 \\  
		\midrule[1pt]
	\end{tabular}
	\caption{Initial results (\%) with Atlas-XL to test its generalizability between NQ and BioASQ.}
	\normalsize
 \label{tab:rq2_init_results}
\end{table}

\section{Corpus-Invariant Tuning} 
Motivated by the observed limitations in the generalization capabilities of retrieval-augmented models, we introduce \emph{Corpus-Invariant Tuning} (CIT) to mitigate memorizing the lexical content of retrieved documents.
%
Specifically, we posit that the generalization difficulties encountered by retrieval-augmented text generation models arises via excessive memorization the documents retrieved from the external corpus by the reader.
In order to achieve higher question-answering accuracy during training, the reader tends to ``hard-code'' a large volume of retrieved documents rather than relying on an improved retriever for a better selection of relevant contexts, as is empirically validated in the document retrieval evaluations of Section 4.5.
This limits the model's ability to generalize because excessive memorization of documents by the reader dictates that when the external corpus is updated or transitions to a different domain, the model faces increased difficulty in adapting and correcting its knowledge compared to learning from scratch.\footnote{Such a phenomenon can be caused by the exposure bias problem as discussed in~\cite{yu2023self}.}
 
\begin{figure}[t]
  \centering
  \includegraphics[width=0.48\textwidth]{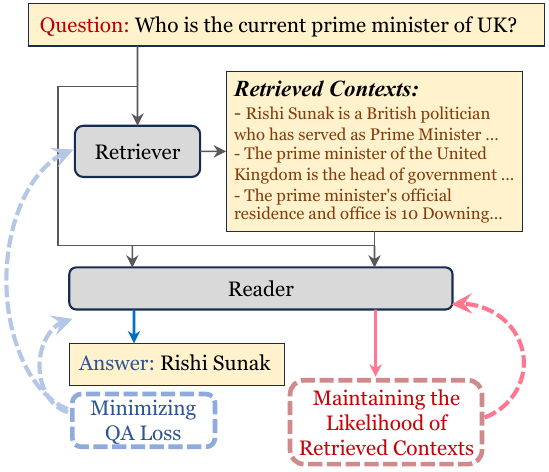}
  \caption{Our proposed Corpus-Invariant Tuning (CIT) Framework. In addition to the existing loss for question answering, we introduce an auxiliary CIT loss to make sure that the reader does not over memorize the retrieved contexts. Specifically, given each batch of QA pairs and the relevant documents retrieved from the corpus, the CIT loss makes sure that the reader's likelihood of these documents does not increase.}\label{fig:framework}
\end{figure}

\paragraph{Validation}
Here we provide an empirical validation on our hypothesis that the degradation of model generalization ability is caused by over-memorization of retrieved knowledge.
We replace the retrieved contexts with ground-truth retrieval results on Wiki-2018, and conduct a stand-alone evaluation with the reader.
We report both the EM score and the overlap rate, i.e., the percentage of incorrectly predicted answers that have overlaps with the ground-truth retrieval results in Wiki-2017.
The results are shown in Table~\ref{tab:validation}.
We can observe that while the models transferred from Wiki-2017 perform slightly worse in terms of EM score, it has a lot more error cases that overlap with the retrieved documents on Wiki-2017.
Such results directly show that the over-memorization of contexts is the primary cause of the degradation of model generalizability.
\begin{table}[htbp]
	\centering
        \small
	\begin{tabular}{cccc}
		\toprule[1pt] 
            
		  \multirow{2}{*}{Dataset} & Training  & EM & Overlap \\
		   ~ &Corpus & Score & Rate\\
		\midrule[1pt]
            
		\multirow{2}{*}{NQ} & Wiki-2017$\rightarrow$2018 &    63.6 & \textbf{76.3}\\
            ~ &Wiki-2018 &     65.2 & 30.2 \\
         \specialrule{0em}{1pt}{1pt}
        \cdashline{1-4}
        \specialrule{0em}{1pt}{1pt}
          \multirow{2}{*}{TriviaQA} & Wiki-2017$\rightarrow$2018 &    78.1 & \textbf{80.8}\\
            ~ &Wiki-2018 &     78.7 & 41.0 \\
		\midrule[1pt]
	\end{tabular}
	\caption{Stand-alone evaluation results (\%) with ground-truth retrieved documents on Wiki-2018. We report both the EM score, and the overlap rate between the incorrectly predicted answers with the retrieval results from the old corpus (Wiki-2017).}
	\normalsize
 \label{tab:validation}
\end{table}
\paragraph{Corpus-Invariant Tuning (CIT)}
To solve this problem, we propose \emph{Corpus-Invariant Tuning} (CIT), a straightforward but effective method to temper the reader's tendency to over-memorize the contents of externally retrieved documents, thereby improving the model's generalization abilities for downstream tasks like OpenQA.
As depicted in Figure~\ref{fig:framework}, the core idea of CIT is to control the reader's memorization (likelihood) of the corpus to be ``invariant'' by introducing an additional loss term that ensures the reader's likelihood of the retrieved documents does not increase during training.
Specifically, for each training QA pair $(x,y)$ and its retrieved document set $\mathcal{C}$, the loss term can be written as
\begin{equation}\label{eqn:KIT}\mathcal{L}_{\textit{CIT}}=\sum_{c\in\mathcal{C}}\|\log p_{\phi}\left(c\right)-\log p_{\phi_{0}}(c)\|^2,
\end{equation}
where $\phi$ and $\phi_0$ denote the current parameters and the original parameters\footnote{$\phi_0$ denotes the initial reader's parameters before training.} of the reader respectively.
We use $p_{\phi}(c)$ to represent the reader's likelihood of a retrieved document $c$.
In our experiments, we adopt the Masked Span Prediction (MSP) probability from the T5 model~\cite{t5} to maintain consistency with the Atlas architecture.
Essentially, we randomly mask out a fixed number of spans of the input sentence and use the model's probability of generating these spans in the correct order as the likelihood.
The overall training objective is a combination of the original loss for question answering $\mathcal{L}_{\textit{QA}}$ and the CIT loss $\mathcal{L}_{\textit{CIT}}$:
\begin{equation}\label{eqn:final_loss}
    \mathcal{L} = \mathcal{L}_{\textit{QA}} + \alpha\cdot\mathcal{L}_{\textit{CIT}},
\end{equation}
where $\alpha$ is a configurable hyper-parameter.

\paragraph{Discussion}
Retrieval-augmented QA models typically maximize answer accuracy as the end-to-end training objective.
However, given the distinct roles of retrievers and readers, there exist two distinct approaches through which this goal can be achieved:
the model can either choose to enhance the retriever to fetch more relevant documents or simply allow the reader to memorize pertinent knowledge.
While both methods can contribute to performance improvements, the former approach increases generalization and CIT biases the model away from rigid memorization by the reader.

\section{Experiments}
\subsection{Data}
Our experiments are conducted on two general-domain OpenQA datasets, NaturalQuestion (NQ) \cite{nq} and TriviaQA \cite{tqa}, and a large cross-domain benchmark RobustQA~\cite{han-etal-2023-robustqa}.
Detailed statistics of these datasets are depicted in Table~\ref{tab:dataset}.
\paragraph{NQ and TriviaQA}
NQ and TriviaQA are the two most widely-used open-domain QA benchmarks, which contain factual question and answer pairs created and annotated based on Wikipedia. 
There are 79,168 and 78,785 QA pairs for training in NQ and TriviaQA respectively.
In our experiments, we conduct both fully supervised training and few-shot training settings for model evaluation.

\paragraph{RobustQA}
RobustQA\footnote{\url{https://github.com/awslabs/robustqa-acl23}} is a large-scale OpenQA evaluation benchmarks specifically designed for evaluating the cross-domain generalization capabilities of OpenQA models.
RobustQA includes 8 distinct domains, each equipped with its own test set and a corresponding list of background documents.
The QA pairs and the documents are adopted and annotated from FiQA,\footnote{\url{https://sites.google.com/view/fiqa/home}} SearchQA~\cite{dunn2017searchqa}, BioASQ~\cite{tsatsaronis2015overview}, and LOTTE~\cite{lotte}.
\begin{table}[hbtp]
	\centering
        \small
	\begin{tabular}{cccc}
		\toprule[1pt] 
            
		  \multirow{2}{*}{Benchmark} & \multirow{2}{*}{Domain} & \# Test & Corpus \\
            ~ & ~ & Questions & Size \\
		\midrule[1pt]
		NQ & Wikipedia & 3,610 & - \\
            \specialrule{0em}{1pt}{1pt}
		\hline
		\specialrule{0em}{1pt}{1pt}
            TriviaQA &  Wikipedia & 11,313 & - \\
            \specialrule{0em}{1pt}{1pt}
		\hline
		\specialrule{0em}{1pt}{1pt}
            \multirow{8}{*}{RobustQA} &  Web Search & 31,760 & 13,791,373 \\
            ~ &  Biomedical & 1,956 & 15,559,026 \\
            ~ &  Finance & 3,669 & 57,638 \\
            ~ &  Lifestyle & 2,214 & 119,461 \\
            ~ &  Recreation & 2,096 & 166,975 \\
            ~ &  Technology & 2,115 & 638,509 \\
            ~ &  Science & 1,426 & 1,694,164 \\
            ~ &  Writing & 2,696 & 199,994 \\
            \midrule[1pt]
	\end{tabular}
	\caption{Detailed statistics of OpenQA evaluation benchmarks used in our experiments.}
	\normalsize
 \label{tab:dataset}
\end{table}
\subsection{Baselines}
We adopt the state-of-the-art retrieval-augmented language model Atlas-XL~\cite{atlas} as our main baseline, which uses a Contriever~\cite{contriever} as the retriever, and a Fusion-in-Decoder (FiD) model~\cite{fid} as the reader.
The primary objective of our experiments is to evaluate whether the baseline model demonstrates improved performance when trained using our proposed CIT loss.
Besides, we also introduce other most recent models Flan-T5~\cite{flan-t5}, RGF~\cite{rgf}, ReAtt~\cite{jiang-etal-2022-retrieval}, FiE+PAQ~\cite{kedia-etal-2022-fie}, and FID-KD~\cite{izacard-grave-2021-leveraging} for comparison.
Our model is labeled as \emph{Atlas-XL}+\emph{CIT} which applies an additional CIT loss to control the knowledge over-memorization of the reader.
\subsection{RQ1: Different versions of the corpus.}

\begin{table*}[hbtp]
	\centering
        \small
	\begin{tabular}{cccccc}
		\midrule[1pt]
		\multirow{2}{*}{\emph{Model}} & \multirow{2}{*}{\emph{Setting}} & \emph{Training} & \emph{Testing} & \multirow{2}{*}{\emph{NQ}} &  \multirow{2}{*}{\emph{TriviaQA}} \\
        ~ & ~ & \emph{Corpus} & \emph{Corpus} & ~ & ~  \\
		\midrule[1pt]
            Atlas-XL~\cite{atlas} &  \emph{Closed Book}  &-&- & 30.2 & 41.6  \\
            \specialrule{0em}{1pt}{1pt}
		\hline
		\specialrule{0em}{1pt}{1pt}
            FiD-KD ~\cite{izacard-grave-2021-leveraging} & \emph{Original} & Wiki-2018  & Wiki-2018 & 54.7 &  67.6\\
            ReAtt~\cite{jiang-etal-2022-retrieval} & \emph{Original} & Wiki-2018  & Wiki-2018 & 54.7 & -\\
            FiE+PAQ~\cite{kedia-etal-2022-fie} & \emph{Original} & Wiki-2018  & Wiki-2018 & 58.4 & 72.6\\
            \specialrule{0em}{1pt}{1pt}
		\hline
		\specialrule{0em}{1pt}{1pt}
		Atlas-XL & \multirow{2}{*}{\emph{Original}} & Wiki-2017  & Wiki-2017 & 58.8 & 75.5  \\
		Atlas-XL + CIT & ~ & Wiki-2017  & Wiki-2017 & 58.9 & 75.5  \\
            \specialrule{0em}{1pt}{1pt}
		\hline
		\specialrule{0em}{1pt}{1pt}
            Atlas-XL & \emph{Zero-shot} & Wiki-2017 & Wiki-2018 & 56.9 & 75.1  \\
		Atlas-XL + CIT & \emph{Transfer} & Wiki-2017 & Wiki-2018 & \textbf{58.6} & \textbf{75.5} \\
            \specialrule{0em}{1pt}{1pt}
		\cdashline{1-6} 
		\specialrule{0em}{1pt}{1pt}
            Atlas-XL & \emph{Full-training} & Wiki-2017$\rightarrow$Wiki-2018 & Wiki-2018 & 59.5 & 76.8 \\
		Atlas-XL + CIT & \emph{Transfer} & Wiki-2017$\rightarrow$Wiki-2018 & Wiki-2018 & \textbf{61.6} & \textbf{77.4} \\
            \midrule[1pt]
	\end{tabular}
	\caption{OpenQA results on the NQ and TriviaQA benchmarks, evaluated with the exact match (EM) score.}
	\normalsize
 \label{tab:rq1_full_results}
\end{table*}
Corresponding to the two research questions proposed in previous sections, we first focus on evaluating our model's ability to generalize across different versions of the external corpus.
We adopt the Wikipedia-domain benchmarks NQ and TriviaQA in our experiments, and test their cross-corpus generalization abilities on different Wikipedia versions.
Similar to the setting in our preliminary experiments presented in Section \ref{sec:pre_expts}, we use Wiki-2017 and Wiki-2018 as our background corpora, and we consider both zero-shot and fully-supervised settings to test the generalization ability of a trained model.
Specifically, we use the following terms to denote different experiment settings:
\begin{itemize}
    \item \emph{Closed Book}: The model is trained and tested without a retriever. The reader is responsible to understand questions and provide answers.
    \item \emph{Original}: Also known as the \emph{Open Book} setting. The most typical experiment setting for retrieval-augmented models, where a retriever retrieves a set of documents from the external corpus, and the reader uses these documents to generate     answer. We use the label \emph{Original} to emphasize the absence of cross-corpus generalization in this setting, providing a baseline for comparison with the following settings to evaluate the model's generalization ability.
    \item \emph{Zero-shot Transfer}: The model is initially trained with the older version of a knowledge corpus, and directly tested with the updated corpus version in a zero-shot manner without any additional fine-tuning.
    \item \emph{Full-training Transfer}: As opposite to the zero-shot transfer setting, after being initially trained with the older version of a knowledge corpus, the model is further fine-tuned on the same training QA pairs with an updated version of the corpus, before being tested with the new corpus.
\end{itemize}

\paragraph{Main Results} 
We conduct generalization experiments in both zero-shot and full-training settings on the NQ and the TriviaQA datasets, and the results are shown in Table~\ref{tab:rq1_full_results}.
In the \emph{Original} setting to train and test the model on the same corpus (Wiki-2017), we can observe that compared with the baseline model Atlas-XL, incorporating a CIT loss to reduce knowledge over-memorization will not diminish the task performance; in fact, it can even slightly enhance it in certain cases (NQ with Wiki-2017).
This is probably because when the reader is discouraged from rigid knowledge memorization, the retriever still has enough room of improvement to retrieve better documents and enhance the performance.
Besides, in both of the \emph{Zero-shot Transfer} and \emph{Full-training Transfer} settings, we can observe that the CIT loss can significantly improve the generalization performance of the model across different versions of a knowledge corpus.
This is likely because when the reader is discouraged from hard-coding knowledge into its parameters, it becomes more receptive to assimilating and utilizing new information from the retriever.
In summary, our proposed CIT loss significantly improves the model's ability to generalize across different versions of external corpus, without compromising the absolute task performance for OpenQA.

\subsection{RQ2: Different domains.}
To address our second research question, in this section, we conduct experiments to evaluate whether our proposed CIT loss can help model better generalize across different knowledge domains. 
We conduct evaluations using the RobustQA benchmark, which encompasses eight diverse domains specifically tailored for OpenQA.
We first assess the model's ability to generalize from a general domain (Wikipedia) to these eight diverse domains.
Subsequently, we evaluate the model's effectiveness in generalizing interchangeably across these eight domains.

\paragraph{From Wikipedia to Specific Domains}
\begin{table*}[t]
	\centering
        \small
	\begin{tabular}{cccccccccc}
		\midrule[1pt]
		\multirow{2}{*}{\emph{Method}} & \multirow{2}{*}{\textbf{\emph{Average}}} & \multirow{2}{*}{\emph{Biomedical}} & \multirow{2}{*}{\emph{Search}} & \multirow{2}{*}{\emph{Finance}} & \multirow{2}{*}{\emph{Life}} & \multirow{2}{*}{\emph{Recreation}} & \multirow{2}{*}{\emph{Technology}} & \multirow{2}{*}{\emph{Science}} & \multirow{2}{*}{\emph{Writing}} \\
        ~  & ~ & ~ & ~ & ~  & ~ & ~ & ~ & ~ & ~  \\
		\midrule[1pt]
            RGF &  23.5  &33.8 & 49.0 & 13.2 & 20.2 & 19.1 & 17.1 & 15.5 & 20.3  \\
            Flan-T5-XL &  32.1 & 43.1 & 70.9 & 14.6& 25.5 & 25.4 & 21.3 & 23.9 & 32.1  \\
            \specialrule{0em}{1pt}{1pt}
		\hline
		\specialrule{0em}{1pt}{1pt}
            Atlas-base &  28.3 & 40.0 & 59.2 & \textbf{15.6} & 23.8 & 22.8 & 19.8 & 18.3 & 27.3  \\
            Atlas-base + CIT & \textbf{30.5} & \textbf{40.3} & \textbf{60.9} & 15.5 & \textbf{26.7} & \textbf{24.9} & \textbf{19.9} & \textbf{23.6} & \textbf{32.3}  \\
            \specialrule{0em}{1pt}{1pt}
  \cdashline{1-10} 
		\specialrule{0em}{1pt}{1pt}
            Atlas-XL &  33.2  & 41.2 & 61.0 & 19.9 & 32.0 & 27.9 & 22.2 &24.8 & 36.7 \\
            Atlas-XL + CIT & \textbf{35.4} & \textbf{43.7} & \textbf{71.5} & \textbf{20.1} & \textbf{33.8} & \textbf{28.1} & \textbf{22.2} & \textbf{26.9} & \textbf{37.0}  \\
            \midrule[1pt]
	\end{tabular}
	\caption{The evaluation results on RobustQA of how well the model can generalize across different domains. All the models are initially fine-tuned on the NQ dataset, and then directly tested on the 8 different domain-specific benchmarks in RobustQA. The performances are evaluated by F1 score to be consistent with the RobustQA paper.}
	\normalsize
 \label{tab:rq2_results}
\end{table*}
We first evaluate how well a model trained with Wikipedia can generalize on the eight specific domains in RobustQA.
Specifically, all the models are first fine-tuned on the NQ dataset with Wiki-2018, and then tested with the eight domain-specific benchmarks.
The results are presented in Table~\ref{tab:rq2_results}.
In general, adding the CIT loss to an Atlas-XL model significantly improves the average F1 score across eight domains in RobustQA, creating a new state-of-the-art among all 3B(XL)-sized models.
Our proposed CIT loss can also help on smaller sizes of models, like Atlas-base, with 2.2\% absolute improvement of the F1 score. 
Within these domains, we can see that domains like \emph{Life} show significant improvement.
This is probably due to their larger overlaps with Wikipedia, which makes it more crucial to avoid over-memorization, so that the old Wikipedia knowledge will not affect generalization on the new domain.
In contrast, domains like \emph{Biomedical} exhibits less improvements.
This is possibly because a \emph{Biomedical} domain KB has a smaller overlap with Wikipedia, thereby reducing the negative impact of knowledge over-memorization on the model's generalizability.

\paragraph{Cross-Domain Generalization}

\begin{figure*}[t]
  \centering
  \includegraphics[width=1.0\linewidth]{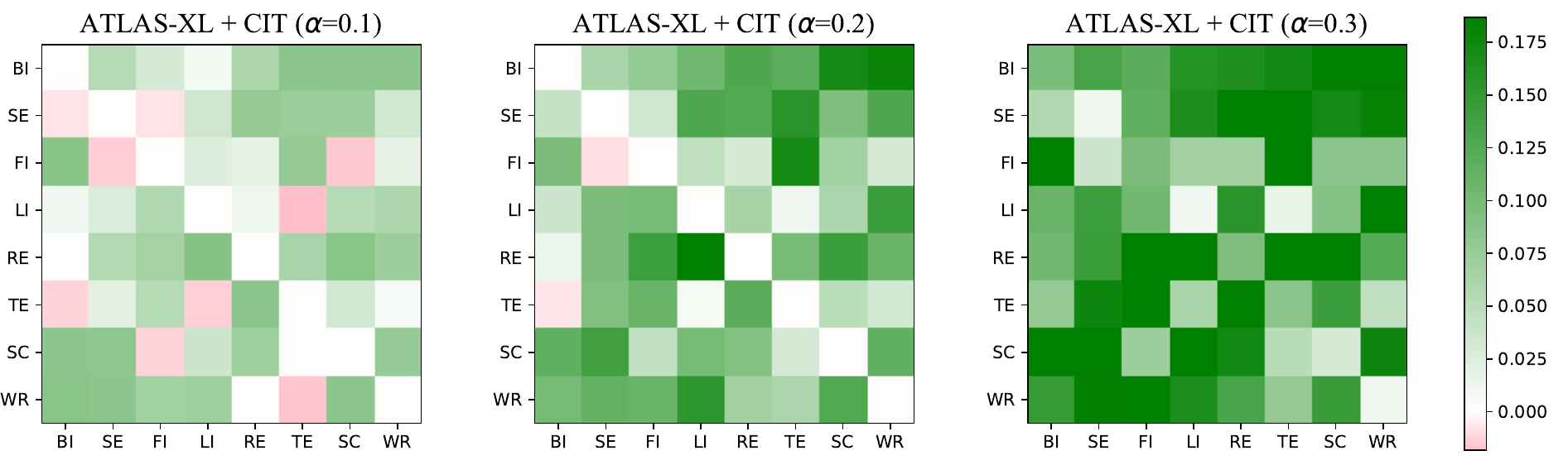}
  \caption{The result heatmaps for cross-domain generalization experiments. Each value in the heatmap represents the absolute improvement (compared with Atlas-XL) of cross-domain relative performance (CRP) defined in Equation~\ref{eqn:crp}. Darker green indicates larger improvements in cross-domain generalization.}\label{fig:heatmap}
\end{figure*}
In addition to testing the model trained from Wikipedia, we also assess the model's capability to generalize between each pair of domains in RobustQA. 
We define the \emph{cross-domain relative performance (CRP)} as the evaluation metric to intuitively characterize how well the model can generalize from a source domain $s$ to a target domain $t$.
Specifically, the $\textit{CRP}(s,t)$ is defined as the ratio of cross-domain performance and intra-domain performance:
\begin{equation}\label{eqn:crp}
    \textit{CRP}(s,t)=\frac{\textit{score}(s,t)}{\textit{score}(t,t)}
\end{equation}
where $\textit{score}(s,t)$ is the performance (F1-score) of training the model in the source domain $s$ and testing the model in the target domain $t$.
In Figure~\ref{fig:heatmap}, we set different CIT strength $\alpha$, and visualize the model's generalizability into a heatmap.
Each heat value $h(s,t)$ stands for the absolute improvements of CRP over the baseline model Atlas-XL:
\begin{equation*}
    h(s,t)=\textit{CRP}_{\alpha}(s,t) - \textit{CRP}_{\textit{Atlas-XL}}(s,t),
\end{equation*}
where darker green indicates larger improvements and darker red indicates larger declines.
We can already observe improvements for most domain pairs with $\alpha=0.1$, and while $\alpha$ reaches to $0.3$, the improvements become much more significant and all domain pairs benefit from CIT in terms of cross-domain generalization.
\subsection{Effect on Retrieval Performance}
\label{sec:retrieval_performance}
The proposed CIT training loss reduces the reader model's memorization tendency, leading to greater reliance on the documents retrieved. This enhanced dependency during training on the retrieved documents appears to enhance the retriever's performance, as seen from the improvements in retrieval performance observed in Table \ref{tab:retrieval_performance} upon integrating CIT loss. Additionally, we measured the coverage of the reader's predicted answer in the retrieved documents for the NQ benchmark, noting an increase in coverage within the top 40 documents from 66.9\% to 69.1\%. This suggests that our proposed CIT training loss leads to an increased reliance by the reader on retrieved documents rather than corpus memorization.

\begin{table}[t]
    \centering

    \begin{tabular}{cccc}
    \midrule[1pt]
       \textit{Model}  &  \textit{R@10}  & \textit{R@20} & \textit{R@40}  \\
      \midrule[1pt] 
       Atlas-XL  & 79.7 & 84.3 & 88.4  \\
       Atlas-XL + CIT & \textbf{85.2} & \textbf{88.9} & \textbf{91.5} \\
       \midrule[1pt]
    \end{tabular}
    \caption{Retrieval performance (Recall@K) on the NQ benchmark.}
    \label{tab:retrieval_performance}
\end{table}


\subsection{Parameter Sensitivity}
\begin{figure}[t]
  \centering
  \includegraphics[width=0.48\textwidth]{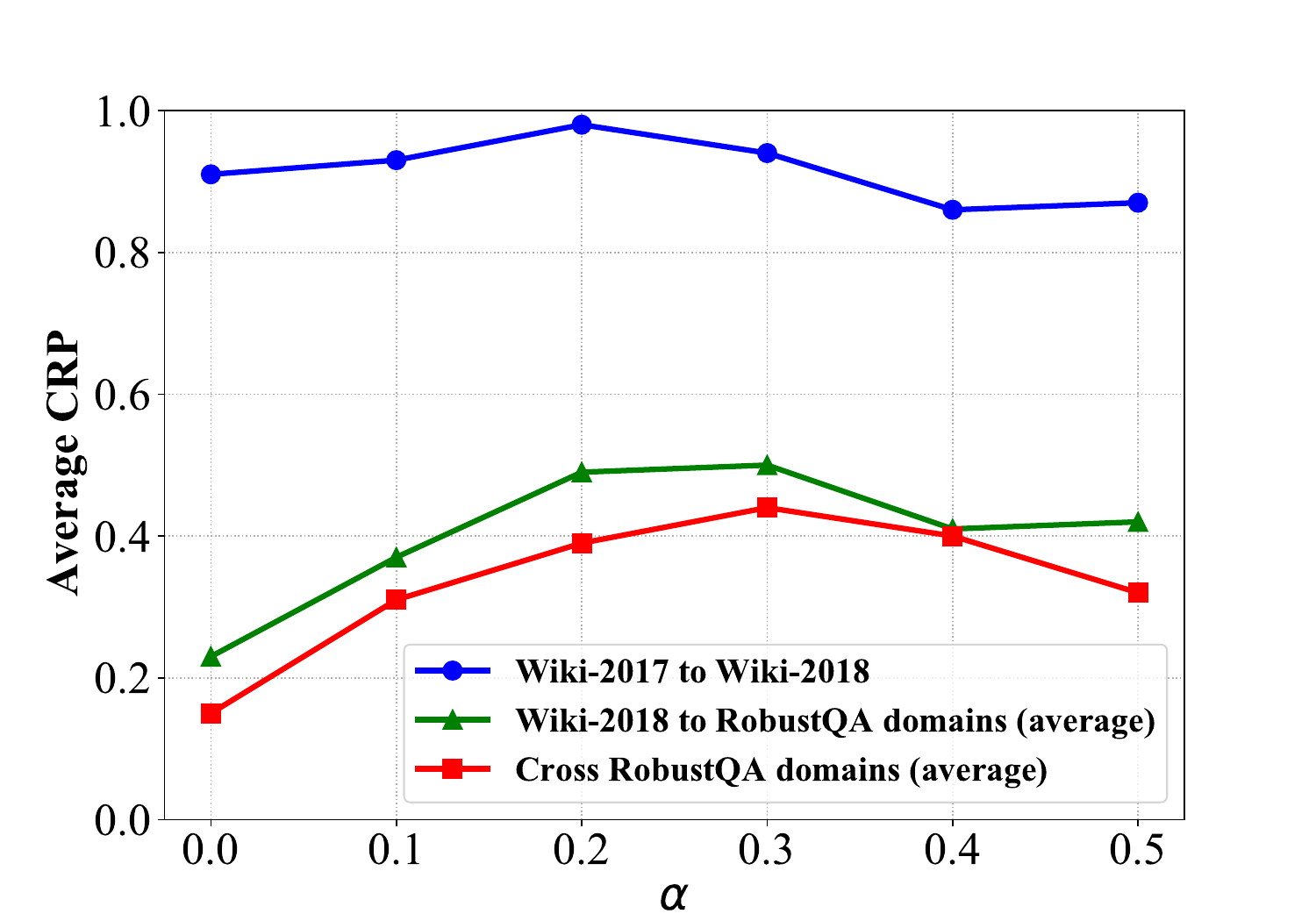}
  \caption{Parameter sensitivity on choices of $\alpha$.}\label{fig:sensitivity}
\end{figure}
We then conduct a more in-depth study on the model sensitivity of the most important hyper-parameter $\alpha$, which controls the strength of the corpus-invariant tuning (as shown in Equation~\ref{eqn:final_loss}).
By choosing different values of $\alpha$, we compute the average cross-domain relative performance between both different corpus versions (RQ1) and different domains (RQ2), and the results are shown in Figure~\ref{fig:sensitivity}.
We can observe that as $\alpha$ becomes larger, the generalization performance initially improves, and then starts to decline after reaching its peak.
This trend is likely because if memorization is controlled too excessively, the reader may neglect memorizing some certain shared global knowledge that is actually beneficial for knowledge generalization.
From Figure~\ref{fig:sensitivity} we can observe that $\alpha=0.2$ is best for generalization across different corpus versions (RQ1), and $\alpha=0.3$ is the best for cross-domain setting (RQ2).
\section{Related Work}
\paragraph{Retrieval-Augmented Text Generation}
Enhancing text generation through the use of retrieved contexts has proven effective in a variety of knowledge-intensive downstream tasks.
The most typical design for retrieval augmentation is to employ a retriever, which is jointly optimized with the reader in an end-to-end manner~\cite{realm, fid,karpukhin-etal-2020-dense, izacard-grave-2021-leveraging, kedia-etal-2022-fie, jiang-etal-2022-retrieval}.
The training methods include iterative training~\cite{iterative_training}, and also EM-based algorithms to treat retrieved documents as hidden variables~\cite{em}.
Retrieval augmentation can also act as plug-in modules~\cite{yu-etal-2023-augmentation}.
In this setting, the retrievers are not jointly trained with the reader, and the retrieved documents are used in an in-context manner~\cite{icralm, raven}.
Recently, Atlas~\cite{atlas} trains and releases a new set of retrieval-augmented models based on the T5 architectures~\cite{t5}, which achieves state-of-the-art performances on OpenQA tasks in few-shot settings.

\paragraph{Open-Domain Question Answering}
The task of OpenQA~\cite{tqa, nq, chen-yih-2020-open} aims at building models to answer questions without background documents.
Given the high demand for external knowledge in this task, the standard approach involves integrating external corpora with a knowledge retriever to supply supporting evidence for answering questions~\cite{fid, DBLP:journals/corr/abs-2006-05244}.
Recently, researchers also focus on new problem settings such as conversational QA~\cite{DBLP:journals/corr/abs-2211-09401}, multi-hop QA~\cite{multihop}, and new knowledge-enhanced solutions like using knowledge graphs~\cite{kg}, and multi-hop reasoning~\cite{reasoning}.

\paragraph{Knowledge Generalization}
Because retrieval-augmented OpenQA models require an external KB to provide supporting documents, it is important to make sure that the model is robust and generalizable across different versions and domains of knowledge.
There are a few previous studies that focus on the generalizability of OpenQA models.
For example, \citet{liu-etal-2022-challenges} focus on the question generalizability of OpenQA models, and conducts a detailed analysis on the generalization performances of current OpenQA models. \citet{gangi2021synthetic, reddy2022towards, reddy2022entity} propose training with synthetic data to improve the robustness of retrieval models in OpenQA settings.
\citet{reconstruction} focuses on the domain adaption problem of OpenQA, and proposes a reconstruction-based auxiliary loss to improve the model's generalizability.
Regarding dataset development, RobustQA~\cite{han-etal-2023-robustqa} creates a new benchmark that involves multiple real-world domains.
However, there are no previous studies that aim to tackle the model's generalization ability on both different corpus and different knowledge domains.
Also, we are the first to tackle this problem in a knowledge memorization perspective, enhancing the model's generalization ability by reducing rigid memorization of the reader modules.

\section{Conclusion}
In this paper, we present \emph{Corpus-Invariant Tuning} (CIT), a simple but effective training strategy to improve the generalization ability of a retrieval-augmented text generation model across different corpus versions and different knowledge domains. 
The main idea of CIT is to mitigate rigid knowledge memorization during training, so that the reader module can easily accept new knowledge in the retrieved documents and adapt to novel unseen domains.
Specifically, we control the reader's likelihood of the retrieved documents during training, to make sure that over-memorization of corpus knowledge is prevented.
Extensive experiments are conducted on multiple OpenQA datasets in both zero-shot and fully-supervised training settings, and the results demonstrate that training the model with the proposed CIT loss significantly improves the model's generalizability across different corpus versions and knowledge domains, without sacrificing the model's inherent performance in its original domain.

\section{Limitations}
Although retrieval-augmented text generation models are effective for many knowledge-intensive tasks, they have an inherent limitation of the large requirement of computational memory.
To ensure time efficiency in the retrieval process, an index of the external corpus, often vast in size, must be pre-constructed.
Another notable limitation of CIT is that the extent of memorization mitigation depends on a hyper-parameter, which is experimented and chosen by humans.
Ideally, the model should be able to automatically determine the best level of memorization mitigation to reach an optimal balance between parametric knowledge and retrieval augmentation.
This is an exciting new research topic, and we will explore this as the future work.

\section*{Acknowledgement}
We thank the anonymous reviewers for their constructive suggestions.
This research is based upon work supported by U.S. DARPA KAIROS Program No. FA8750-19-2-1004. The views and conclusions contained herein are those of the authors and should not be interpreted as necessarily representing the official policies, either expressed or implied, of DARPA, or the U.S. Government. The U.S. Government is authorized to reproduce and distribute reprints for governmental purposes notwithstanding any copyright annotation therein.

\bibliography{anthology,custom}

\begin{thebibliography}{38}
\expandafter\ifx\csname natexlab\endcsname\relax\def\natexlab#1{#1}\fi

\bibitem[{Chen and Yih(2020)}]{chen-yih-2020-open}
Danqi Chen and Wen-tau Yih. 2020.
\newblock \href {https://doi.org/10.18653/v1/2020.acl-tutorials.8} {Open-domain question answering}.
\newblock In \emph{Proceedings of the 58th Annual Meeting of the Association for Computational Linguistics: Tutorial Abstracts}, pages 34--37, Online. Association for Computational Linguistics.

\bibitem[{Chung et~al.(2022)Chung, Hou, Longpre, Zoph, Tay, Fedus, Li, Wang, Dehghani, Brahma, Webson, Gu, Dai, Suzgun, Chen, Chowdhery, Narang, Mishra, Yu, Zhao, Huang, Dai, Yu, Petrov, Chi, Dean, Devlin, Roberts, Zhou, Le, and Wei}]{flan-t5}
Hyung~Won Chung, Le~Hou, Shayne Longpre, Barret Zoph, Yi~Tay, William Fedus, Eric Li, Xuezhi Wang, Mostafa Dehghani, Siddhartha Brahma, Albert Webson, Shixiang~Shane Gu, Zhuyun Dai, Mirac Suzgun, Xinyun Chen, Aakanksha Chowdhery, Sharan Narang, Gaurav Mishra, Adams Yu, Vincent~Y. Zhao, Yanping Huang, Andrew~M. Dai, Hongkun Yu, Slav Petrov, Ed~H. Chi, Jeff Dean, Jacob Devlin, Adam Roberts, Denny Zhou, Quoc~V. Le, and Jason Wei. 2022.
\newblock \href {https://doi.org/10.48550/ARXIV.2210.11416} {Scaling instruction-finetuned language models}.
\newblock \emph{CoRR}, abs/2210.11416.

\bibitem[{Dunn et~al.(2017)Dunn, Sagun, Higgins, Guney, Cirik, and Cho}]{dunn2017searchqa}
Matthew Dunn, Levent Sagun, Mike Higgins, V.~Ugur Guney, Volkan Cirik, and Kyunghyun Cho. 2017.
\newblock \href {http://arxiv.org/abs/1704.05179} {Searchqa: A new q\&a dataset augmented with context from a search engine}.

\bibitem[{Fang et~al.(2022)Fang, Hung, Huang, and Chen}]{DBLP:journals/corr/abs-2211-09401}
Hung{-}Chieh Fang, Kuo{-}Han Hung, Chao{-}Wei Huang, and Yun{-}Nung Chen. 2022.
\newblock \href {https://doi.org/10.48550/ARXIV.2211.09401} {Open-domain conversational question answering with historical answers}.
\newblock \emph{CoRR}, abs/2211.09401.

\bibitem[{Gangi~Reddy et~al.(2021)Gangi~Reddy, Iyer, Sultan, Zhang, Sil, Castelli, Florian, and Roukos}]{gangi2021synthetic}
Revanth Gangi~Reddy, Bhavani Iyer, Md~Arafat Sultan, Rong Zhang, Avirup Sil, Vittorio Castelli, Radu Florian, and Salim Roukos. 2021.
\newblock Synthetic target domain supervision for open retrieval qa.
\newblock In \emph{Proceedings of the 44th International ACM SIGIR Conference on Research and Development in Information Retrieval}, pages 1793--1797.

\bibitem[{Guu et~al.(2020)Guu, Lee, Tung, Pasupat, and Chang}]{realm}
Kelvin Guu, Kenton Lee, Zora Tung, Panupong Pasupat, and Ming{-}Wei Chang. 2020.
\newblock \href {http://arxiv.org/abs/2002.08909} {{REALM:} retrieval-augmented language model pre-training}.
\newblock \emph{CoRR}, abs/2002.08909.

\bibitem[{Han et~al.(2023)Han, Qi, Zhang, Liu, Burger, Wang, Huang, Xiang, and Roth}]{han-etal-2023-robustqa}
Rujun Han, Peng Qi, Yuhao Zhang, Lan Liu, Juliette Burger, William~Yang Wang, Zhiheng Huang, Bing Xiang, and Dan Roth. 2023.
\newblock \href {https://doi.org/10.18653/v1/2023.findings-acl.263} {{R}obust{QA}: Benchmarking the robustness of domain adaptation for open-domain question answering}.
\newblock In \emph{Findings of the Association for Computational Linguistics: ACL 2023}, pages 4294--4311, Toronto, Canada. Association for Computational Linguistics.

\bibitem[{Hu et~al.(2022)Hu, Xu, Yu, Wang, Yang, Zhu, Chang, and Sun}]{reasoning}
Ziniu Hu, Yichong Xu, Wenhao Yu, Shuohang Wang, Ziyi Yang, Chenguang Zhu, Kai{-}Wei Chang, and Yizhou Sun. 2022.
\newblock \href {https://doi.org/10.18653/V1/2022.EMNLP-MAIN.650} {Empowering language models with knowledge graph reasoning for open-domain question answering}.
\newblock In \emph{Proceedings of the 2022 Conference on Empirical Methods in Natural Language Processing, {EMNLP} 2022, Abu Dhabi, United Arab Emirates, December 7-11, 2022}, pages 9562--9581. Association for Computational Linguistics.

\bibitem[{Huang et~al.(2023)Huang, Ping, Xu, Shoeybi, Chang, and Catanzaro}]{raven}
Jie Huang, Wei Ping, Peng Xu, Mohammad Shoeybi, Kevin~Chen{-}Chuan Chang, and Bryan Catanzaro. 2023.
\newblock \href {https://doi.org/10.48550/ARXIV.2308.07922} {{RAVEN:} in-context learning with retrieval augmented encoder-decoder language models}.
\newblock \emph{CoRR}, abs/2308.07922.

\bibitem[{Izacard et~al.(2022)Izacard, Caron, Hosseini, Riedel, Bojanowski, Joulin, and Grave}]{contriever}
Gautier Izacard, Mathilde Caron, Lucas Hosseini, Sebastian Riedel, Piotr Bojanowski, Armand Joulin, and Edouard Grave. 2022.
\newblock \href {https://openreview.net/forum?id=jKN1pXi7b0} {Unsupervised dense information retrieval with contrastive learning}.
\newblock \emph{Trans. Mach. Learn. Res.}, 2022.

\bibitem[{Izacard and Grave(2021{\natexlab{a}})}]{iterative_training}
Gautier Izacard and Edouard Grave. 2021{\natexlab{a}}.
\newblock \href {https://openreview.net/forum?id=NTEz-6wysdb} {Distilling knowledge from reader to retriever for question answering}.
\newblock In \emph{9th International Conference on Learning Representations, {ICLR} 2021, Virtual Event, Austria, May 3-7, 2021}. OpenReview.net.

\bibitem[{Izacard and Grave(2021{\natexlab{b}})}]{izacard-grave-2021-leveraging}
Gautier Izacard and Edouard Grave. 2021{\natexlab{b}}.
\newblock \href {https://doi.org/10.18653/v1/2021.eacl-main.74} {Leveraging passage retrieval with generative models for open domain question answering}.
\newblock In \emph{Proceedings of the 16th Conference of the European Chapter of the Association for Computational Linguistics: Main Volume}, pages 874--880, Online. Association for Computational Linguistics.

\bibitem[{Izacard et~al.(2023)Izacard, Lewis, Lomeli, Hosseini, Petroni, Schick, Dwivedi{-}Yu, Joulin, Riedel, and Grave}]{atlas}
Gautier Izacard, Patrick S.~H. Lewis, Maria Lomeli, Lucas Hosseini, Fabio Petroni, Timo Schick, Jane Dwivedi{-}Yu, Armand Joulin, Sebastian Riedel, and Edouard Grave. 2023.
\newblock \href {http://jmlr.org/papers/v24/23-0037.html} {Atlas: Few-shot learning with retrieval augmented language models}.
\newblock \emph{J. Mach. Learn. Res.}, 24:251:1--251:43.

\bibitem[{Izacard et~al.(2020)Izacard, Petroni, Hosseini, Cao, Riedel, and Grave}]{fid}
Gautier Izacard, Fabio Petroni, Lucas Hosseini, Nicola~De Cao, Sebastian Riedel, and Edouard Grave. 2020.
\newblock \href {http://arxiv.org/abs/2012.15156} {A memory efficient baseline for open domain question answering}.
\newblock \emph{CoRR}, abs/2012.15156.

\bibitem[{Jiang et~al.(2022)Jiang, Gao, Wang, Araki, Ding, Callan, and Neubig}]{jiang-etal-2022-retrieval}
Zhengbao Jiang, Luyu Gao, Zhiruo Wang, Jun Araki, Haibo Ding, Jamie Callan, and Graham Neubig. 2022.
\newblock \href {https://doi.org/10.18653/v1/2022.emnlp-main.149} {Retrieval as attention: End-to-end learning of retrieval and reading within a single transformer}.
\newblock In \emph{Proceedings of the 2022 Conference on Empirical Methods in Natural Language Processing}, pages 2336--2349, Abu Dhabi, United Arab Emirates. Association for Computational Linguistics.

\bibitem[{Joshi et~al.(2017)Joshi, Choi, Weld, and Zettlemoyer}]{tqa}
Mandar Joshi, Eunsol Choi, Daniel Weld, and Luke Zettlemoyer. 2017.
\newblock \href {https://doi.org/10.18653/v1/P17-1147} {{T}rivia{QA}: A large scale distantly supervised challenge dataset for reading comprehension}.
\newblock In \emph{Proceedings of the 55th Annual Meeting of the Association for Computational Linguistics (Volume 1: Long Papers)}, pages 1601--1611, Vancouver, Canada. Association for Computational Linguistics.

\bibitem[{Karpukhin et~al.(2020)Karpukhin, Oguz, Min, Lewis, Wu, Edunov, Chen, and Yih}]{karpukhin-etal-2020-dense}
Vladimir Karpukhin, Barlas Oguz, Sewon Min, Patrick Lewis, Ledell Wu, Sergey Edunov, Danqi Chen, and Wen-tau Yih. 2020.
\newblock \href {https://doi.org/10.18653/v1/2020.emnlp-main.550} {Dense passage retrieval for open-domain question answering}.
\newblock In \emph{Proceedings of the 2020 Conference on Empirical Methods in Natural Language Processing (EMNLP)}, pages 6769--6781, Online. Association for Computational Linguistics.

\bibitem[{Kasai et~al.(2022)Kasai, Sakaguchi, Takahashi, Bras, Asai, Yu, Radev, Smith, Choi, and Inui}]{kasai2022realtime}
Jungo Kasai, Keisuke Sakaguchi, Yoichi Takahashi, Ronan~Le Bras, Akari Asai, Xinyan Yu, Dragomir Radev, Noah~A Smith, Yejin Choi, and Kentaro Inui. 2022.
\newblock Realtime qa: What's the answer right now?
\newblock \emph{arXiv preprint arXiv:2207.13332}.

\bibitem[{Kedia et~al.(2022)Kedia, Zaidi, and Lee}]{kedia-etal-2022-fie}
Akhil Kedia, Mohd~Abbas Zaidi, and Haejun Lee. 2022.
\newblock \href {https://doi.org/10.18653/v1/2022.emnlp-main.285} {{F}i{E}: Building a global probability space by leveraging early fusion in encoder for open-domain question answering}.
\newblock In \emph{Proceedings of the 2022 Conference on Empirical Methods in Natural Language Processing}, pages 4246--4260, Abu Dhabi, United Arab Emirates. Association for Computational Linguistics.

\bibitem[{Kwiatkowski et~al.(2019)Kwiatkowski, Palomaki, Redfield, Collins, Parikh, Alberti, Epstein, Polosukhin, Devlin, Lee, Toutanova, Jones, Kelcey, Chang, Dai, Uszkoreit, Le, and Petrov}]{nq}
Tom Kwiatkowski, Jennimaria Palomaki, Olivia Redfield, Michael Collins, Ankur~P. Parikh, Chris Alberti, Danielle Epstein, Illia Polosukhin, Jacob Devlin, Kenton Lee, Kristina Toutanova, Llion Jones, Matthew Kelcey, Ming{-}Wei Chang, Andrew~M. Dai, Jakob Uszkoreit, Quoc Le, and Slav Petrov. 2019.
\newblock \href {https://doi.org/10.1162/TACL\_A\_00276} {Natural questions: a benchmark for question answering research}.
\newblock \emph{Trans. Assoc. Comput. Linguistics}, 7:452--466.

\bibitem[{Liu et~al.(2022)Liu, Lewis, Riedel, and Stenetorp}]{liu-etal-2022-challenges}
Linqing Liu, Patrick Lewis, Sebastian Riedel, and Pontus Stenetorp. 2022.
\newblock \href {https://doi.org/10.18653/v1/2022.findings-naacl.155} {Challenges in generalization in open domain question answering}.
\newblock In \emph{Findings of the Association for Computational Linguistics: NAACL 2022}, pages 2014--2029, Seattle, United States. Association for Computational Linguistics.

\bibitem[{Oduro{-}Afriyie and Jamil(2023)}]{kg}
Joel Oduro{-}Afriyie and Hasan Jamil. 2023.
\newblock \href {https://doi.org/10.1007/978-3-031-42935-4\_6} {Knowledge graph enabled open-domain conversational question answering}.
\newblock In \emph{Flexible Query Answering Systems - 15th International Conference, {FQAS} 2023, Mallorca, Spain, September 5-7, 2023, Proceedings}, volume 14113 of \emph{Lecture Notes in Computer Science}, pages 63--76. Springer.

\bibitem[{Paranjape et~al.(2022)Paranjape, Lamm, and Tenney}]{rgf}
Bhargavi Paranjape, Matthew Lamm, and Ian Tenney. 2022.
\newblock \href {https://doi.org/10.18653/v1/2022.acl-long.117} {Retrieval-guided counterfactual generation for {QA}}.
\newblock In \emph{Proceedings of the 60th Annual Meeting of the Association for Computational Linguistics (Volume 1: Long Papers)}, pages 1670--1686, Dublin, Ireland. Association for Computational Linguistics.

\bibitem[{Raffel et~al.(2020)Raffel, Shazeer, Roberts, Lee, Narang, Matena, Zhou, Li, and Liu}]{t5}
Colin Raffel, Noam Shazeer, Adam Roberts, Katherine Lee, Sharan Narang, Michael Matena, Yanqi Zhou, Wei Li, and Peter~J. Liu. 2020.
\newblock \href {http://jmlr.org/papers/v21/20-074.html} {Exploring the limits of transfer learning with a unified text-to-text transformer}.
\newblock \emph{J. Mach. Learn. Res.}, 21:140:1--140:67.

\bibitem[{Ram et~al.(2023)Ram, Levine, Dalmedigos, Muhlgay, Shashua, Leyton{-}Brown, and Shoham}]{icralm}
Ori Ram, Yoav Levine, Itay Dalmedigos, Dor Muhlgay, Amnon Shashua, Kevin Leyton{-}Brown, and Yoav Shoham. 2023.
\newblock \href {https://doi.org/10.48550/ARXIV.2302.00083} {In-context retrieval-augmented language models}.
\newblock \emph{CoRR}, abs/2302.00083.

\bibitem[{Reddy et~al.(2022{\natexlab{a}})Reddy, Sultan, Franz, Sil, and Ji}]{reddy2022entity}
Revanth~Gangi Reddy, Md~Arafat Sultan, Martin Franz, Avirup Sil, and Heng Ji. 2022{\natexlab{a}}.
\newblock Entity-conditioned question generation for robust attention distribution in neural information retrieval.
\newblock In \emph{Proceedings of the 45th International ACM SIGIR Conference on Research and Development in Information Retrieval}, pages 2462--2466.

\bibitem[{Reddy et~al.(2022{\natexlab{b}})Reddy, Yadav, Sultan, Franz, Castelli, Ji, and Sil}]{reddy2022towards}
Revanth~Gangi Reddy, Vikas Yadav, Md~Arafat Sultan, Martin Franz, Vittorio Castelli, Heng Ji, and Avirup Sil. 2022{\natexlab{b}}.
\newblock Towards robust neural retrieval with source domain synthetic pre-finetuning.
\newblock In \emph{Proceedings of the 29th International Conference on Computational Linguistics}, pages 1065--1070.

\bibitem[{Roberts et~al.(2020)Roberts, Raffel, and Shazeer}]{roberts2020much}
Adam Roberts, Colin Raffel, and Noam Shazeer. 2020.
\newblock How much knowledge can you pack into the parameters of a language model?
\newblock In \emph{Proceedings of the 2020 Conference on Empirical Methods in Natural Language Processing (EMNLP)}, pages 5418--5426.

\bibitem[{Sachan et~al.(2021)Sachan, Reddy, Hamilton, Dyer, and Yogatama}]{em}
Devendra~Singh Sachan, Siva Reddy, William~L. Hamilton, Chris Dyer, and Dani Yogatama. 2021.
\newblock \href {https://proceedings.neurips.cc/paper/2021/hash/da3fde159d754a2555eaa198d2d105b2-Abstract.html} {End-to-end training of multi-document reader and retriever for open-domain question answering}.
\newblock In \emph{Advances in Neural Information Processing Systems 34: Annual Conference on Neural Information Processing Systems 2021, NeurIPS 2021, December 6-14, 2021, virtual}, pages 25968--25981.

\bibitem[{Santhanam et~al.(2022)Santhanam, Khattab, Saad-Falcon, Potts, and Zaharia}]{lotte}
Keshav Santhanam, Omar Khattab, Jon Saad-Falcon, Christopher Potts, and Matei Zaharia. 2022.
\newblock \href {https://doi.org/10.18653/v1/2022.naacl-main.272} {{C}ol{BERT}v2: Effective and efficient retrieval via lightweight late interaction}.
\newblock In \emph{Proceedings of the 2022 Conference of the North American Chapter of the Association for Computational Linguistics: Human Language Technologies}, pages 3715--3734, Seattle, United States. Association for Computational Linguistics.

\bibitem[{Siriwardhana et~al.(2023)Siriwardhana, Weerasekera, Kaluarachchi, Wen, Rana, and Nanayakkara}]{reconstruction}
Shamane Siriwardhana, Rivindu Weerasekera, Tharindu Kaluarachchi, Elliott Wen, Rajib Rana, and Suranga Nanayakkara. 2023.
\newblock \href {https://transacl.org/ojs/index.php/tacl/article/view/4029} {Improving the domain adaptation of retrieval augmented generation {(RAG)} models for open domain question answering}.
\newblock \emph{Trans. Assoc. Comput. Linguistics}, 11:1--17.

\bibitem[{Touvron et~al.(2023)Touvron, Martin, Stone, Albert, Almahairi, Babaei, Bashlykov, Batra, Bhargava, Bhosale et~al.}]{touvron2023llama}
Hugo Touvron, Louis Martin, Kevin Stone, Peter Albert, Amjad Almahairi, Yasmine Babaei, Nikolay Bashlykov, Soumya Batra, Prajjwal Bhargava, Shruti Bhosale, et~al. 2023.
\newblock Llama 2: Open foundation and fine-tuned chat models.
\newblock \emph{arXiv preprint arXiv:2307.09288}.

\bibitem[{Tsatsaronis et~al.(2015)Tsatsaronis, Balikas, Malakasiotis, Partalas, Zschunke, Alvers, Weissenborn, Krithara, Petridis, Polychronopoulos et~al.}]{tsatsaronis2015overview}
George Tsatsaronis, Georgios Balikas, Prodromos Malakasiotis, Ioannis Partalas, Matthias Zschunke, Michael~R Alvers, Dirk Weissenborn, Anastasia Krithara, Sergios Petridis, Dimitris Polychronopoulos, et~al. 2015.
\newblock An overview of the bioasq large-scale biomedical semantic indexing and question answering competition.
\newblock \emph{BMC bioinformatics}, 16(1):1--28.

\bibitem[{Vu et~al.(2023)Vu, Iyyer, Wang, Constant, Wei, Wei, Tar, Sung, Zhou, Le et~al.}]{vu2023freshllms}
Tu~Vu, Mohit Iyyer, Xuezhi Wang, Noah Constant, Jerry Wei, Jason Wei, Chris Tar, Yun-Hsuan Sung, Denny Zhou, Quoc Le, et~al. 2023.
\newblock Freshllms: Refreshing large language models with search engine augmentation.
\newblock \emph{arXiv preprint arXiv:2310.03214}.

\bibitem[{Xiong et~al.(2021)Xiong, Li, Iyer, Du, Lewis, Wang, Mehdad, Yih, Riedel, Kiela, and Oguz}]{multihop}
Wenhan Xiong, Xiang~Lorraine Li, Srini Iyer, Jingfei Du, Patrick S.~H. Lewis, William~Yang Wang, Yashar Mehdad, Scott Yih, Sebastian Riedel, Douwe Kiela, and Barlas Oguz. 2021.
\newblock \href {https://openreview.net/forum?id=EMHoBG0avc1} {Answering complex open-domain questions with multi-hop dense retrieval}.
\newblock In \emph{9th International Conference on Learning Representations, {ICLR} 2021, Virtual Event, Austria, May 3-7, 2021}. OpenReview.net.

\bibitem[{Yu and Ji(2023)}]{yu2023self}
Pengfei Yu and Heng Ji. 2023.
\newblock \href {http://arxiv.org/abs/2305.18582} {Self information update for large language models through mitigating exposure bias}.

\bibitem[{Yu et~al.(2023)Yu, Xiong, Yu, and Liu}]{yu-etal-2023-augmentation}
Zichun Yu, Chenyan Xiong, Shi Yu, and Zhiyuan Liu. 2023.
\newblock \href {https://doi.org/10.18653/v1/2023.acl-long.136} {Augmentation-adapted retriever improves generalization of language models as generic plug-in}.
\newblock In \emph{Proceedings of the 61st Annual Meeting of the Association for Computational Linguistics (Volume 1: Long Papers)}, pages 2421--2436, Toronto, Canada. Association for Computational Linguistics.

\bibitem[{Zhou et~al.(2020)Zhou, Shi, Huang, and Zhu}]{DBLP:journals/corr/abs-2006-05244}
Mantong Zhou, Zhouxing Shi, Minlie Huang, and Xiaoyan Zhu. 2020.
\newblock \href {http://arxiv.org/abs/2006.05244} {Knowledge-aided open-domain question answering}.
\newblock \emph{CoRR}, abs/2006.05244.

\end{thebibliography}
\appendix

\newpage
\section{Implementation Details}
\label{sec:appendix}
\paragraph{Experiment Details}
We train our model on 4 NVIDIA A100
GPUs with 80 GB memory, and the total training
time is about 3.5 hours for the model to converge on an OpenQA dataset.
During training, the index of the external corpus is pre-computed and equally sharded across all 4 GPUs. We adopt distributed data parallelism to make a copy of the model on each GPU and the data batch is splitted to the 4 devices.

\paragraph{Hyper-Parameter Settings}
Detailed hyper-parameter search range and choices are shown in Table~\ref{tab:hyper-parameter-range} and Table~\ref{tab:hyper-parameter} respectively.
The choices are made by grid search.
\begin{table}[htbp]
	\centering
    \small
	\begin{tabular}{cc}
		\toprule[1pt] 
		Hyper-parameters &  Searching Range\\
		\midrule[1pt]
        Maximum Length of FiD & [384, 512, 768]\\
        \# Retrieved Contexts  & [20, 30, 40]\\
		Generation Length&  [10, 15, 20, 25, 30]\\
		Masked Percentage for CIT &  [0.1, 0.15, 0.2] \\
		Strength of CIT $\alpha$ & [0.1, 0.2, 0.3, 0.4, 0.5, 0.6] \\
		\specialrule{0em}{1pt}{1pt}
		\hline
		\specialrule{0em}{1pt}{1pt}
		Learning Rate  & [1e-5, 2e-5, 3e-5, 4e-5, 5e-5]\\
		Batch Size & [4, 8, 12, 16]\\
		
		Maximum Training Steps & [500, 1,000, 1,500] \\
		Warm-up Steps & [50, 100, 150] \\
            Weight Decay & [1e-2, 1e-3, 1e-4] \\
            Retriever Dropout & [0.1, 0.2, 0.3] \\
            Reader Dropout & [0.1, 0.2, 0.3] \\
		\midrule[1pt]
	\end{tabular}
	\normalsize
	\caption{Detailed hyper-parameter  search ranges for Corpus-Invariant Tuning.}
	\label{tab:hyper-parameter-range}
\end{table}
\begin{table}[htbp]
	\centering
    \small
	\begin{tabular}{cc}
		\toprule[1pt] 
		Hyper-parameters & Values \\
		\midrule[1pt]
        Maximum Length of FiD & 512 \\
        \# Retrieved Contexts & 40 \\
		Generation Length & 20 \\
		Masked Percentage of CIT & 0.15  \\
		Strength of CIT $\alpha$ & 0.2 (RQ1); 0.3 (RQ2) \\
		\specialrule{0em}{1pt}{1pt}
		\hline
		\specialrule{0em}{1pt}{1pt}
		Learning Rate & 4e-5 \\
		Batch Size & 8 \\
		
		Maximum Training Steps & 500 \\
		Warm-up Steps & 50 \\
            Weight Decay & 1e-2 \\
            Retriever Dropout & 0.1  \\
            Reader Dropout & 0.1 \\
		\midrule[1pt]
	\end{tabular}
	\normalsize
	\caption{Detailed hyper-parameter  configurations for Corpus-Invariant Tuning.}
	\label{tab:hyper-parameter}
\end{table}

\section{Case Study on Decoder-Only Models}
The main idea of CIT is to control knowledge over-memorization to improve model's generalization ability among different corpora and different domains. Such a training strategy is not only effective for retrieval-augmented encoder-decoder models like Atlas, but can theoretically also apply to larger-scale auto-regressive foundation models.
Therefore, we also conduct preliminary experiments with LLaMA-2-7b~\cite{touvron2023llama} and Contriever as a case study to evaluate the robustness and ubiquity of our model among different language model architectures.
Specifically, we freeze the retriever, and use the retrieved documents as the input prompt to generate the answer for each question.
While maximizing the probabilities of the correct answers, we apply CIT in a similar way of maintaining the direct log-likelihood of these retrieved contexts $\mathcal{C}$:
\begin{equation*}
\mathcal{L}_{\textit{CIT}}=\sum_{c\in\mathcal{C}}\|\log p'_{\phi}\left(c\right)-\log p'_{\phi_{0}}(c)\|^2.
\end{equation*}
Different from the masked span prediction probability $p_{\phi}$ in Equation~\eqref{eqn:KIT}, the $p'_{\phi}$ here represents the language modeling probability of the entire passage $c$. 
Denoting the $i$-th word in $c$ as $c_i$, then $p'_{\phi}$ can be formulated by
\begin{equation*}
p'_{\phi}(c) = \prod_{i=1}^{N-1}p\left(c_{i+1} \mid c_{1:i}\right).
\end{equation*}
As shown in Table~\ref{tab:llama}, we can observe that with auto-regressive language models like LLaMA, our proposed CIT can still achieves considerable improvements of the model's generalization ability.
\begin{table}[htbp]
	\centering
        \small
	\begin{tabular}{ccc}
		\toprule[1pt] 
            
		  Dataset & Model  & EM Score  \\
		\midrule[1pt]
		\multirow{2}{*}{NQ} & LLaMA2-7b &    54.2 \\
            ~ &LLaMA2-7b + CIT &     57.7  \\
         \specialrule{0em}{1pt}{1pt}
        \cdashline{1-3}
        \specialrule{0em}{1pt}{1pt}
          \multirow{2}{*}{TriviaQA} & LLaMA2-7b &    69.9 \\
            ~ &LLaMA2-7b + CIT &     71.4 \\
		\midrule[1pt]
	\end{tabular}
	\caption{Evaluation results of applying the idea of CIT on the LLaMA2-7b model with a fixed retriever (Contriever). The experiments are conducted in a zero-shot transfer setting between Wiki-2017 and Wiki-2018.}
	\normalsize
 \label{tab:llama}
\end{table}
\end{document}